\documentclass[letterpaper, 10 pt, conference]{ieeeconf}  

\IEEEoverridecommandlockouts                              

\pdfoutput=1
\usepackage{epsfig} 
\usepackage{amsmath} 
\usepackage{amssymb}  
\usepackage{booktabs}
\usepackage{array}
\usepackage{float}
\usepackage{multirow}
\usepackage{subfigure}
\usepackage{url}
\usepackage{cite}
\usepackage{algorithm}  
\usepackage{algorithmic}

\title{\LARGE \bf
	Generic Probabilistic Interactive Situation Recognition and Prediction: \\
	From Virtual to Real 
}

\author{Jiachen Li, Hengbo Ma, 
	Wei Zhan 
	and Masayoshi Tomizuka
	\thanks{J. Li, H. Ma, W. Zhan and M. Tomizuka are with the Department of Mechanical Engineering, 
		University of California, Berkeley, CA 94720, USA
		(e-mail: {\tt\small jiachen\_li, hengbo\_ma, wzhan, tomizuka@berkeley.edu})}}
	
\begin{document}
\maketitle
\thispagestyle{empty}
\pagestyle{empty}

\begin{abstract}
Accurate and robust recognition and prediction of traffic situation plays an important role in autonomous driving, which is a prerequisite for risk assessment and effective decision making. Although there exist a lot of works dealing with modeling driver behavior of a single object, it remains a challenge to make predictions for multiple highly interactive agents that react to each other simultaneously. 
In this work, we propose a generic probabilistic hierarchical recognition and prediction framework which employs a two-layer Hidden Markov Model (TLHMM) to obtain the distribution of potential situations and a learning-based dynamic scene evolution model to sample a group of future trajectories. Instead of predicting motions of a single entity, we propose to get the joint distribution by modeling multiple interactive agents as a whole system. 
Moreover, due to the decoupling property of the layered structure, our model is suitable for knowledge transfer from simulation to real world applications as well as among different traffic scenarios, which can reduce the computational efforts of training and the demand for a large data amount. A case study of highway ramp merging scenario is demonstrated to verify the effectiveness and accuracy of the proposed framework. 
\end{abstract}

\section{INTRODUCTION}

Accurate and efficient recognition and prediction of future traffic scene evolution plays a significant role in autonomous driving which is a prerequisite for risk assessment, decision making and high-quality motion planning. Although a lot of research efforts have been devoted to the driver behavior recognition and prediction, most of them only focused on a single entity \cite{A3,C1,C3,C2,CPN}  while the information on surrounding vehicle is obtained from onboard sensor measurement and utilized as prior knowledge when making predictions. In recent years, more attention has been paid to model the interaction among multiple agents. 
A review on motion prediction and risk assessment was provided in \cite{A7} where the behavior models are classified into three categories: physics-based, maneuver-based and interaction-aware models. While the first two types are crucial for motion planning and control purpose, the models involving more interaction factors are important for the threat warning system and the decision making module in the Advanced Driver Assistant System (ADAS) as well as in fully autonomous vehicles \cite{A6,A8,YH_prediction,A5}.

There remain several limitations and challenges when modeling human-like interactions: 1) It is hard to collect real-world driving data containing strong interactions among multiple traffic participants; 2) There is no ground truth for the driver intention since the human desire can vary from time to time. 
Therefore, it is desired to  have an effective and robust recognition and prediction framework which takes into account uncertainty in human behavior and does not require a large amount of data.
\begin{figure}
	\centering
	\includegraphics[width=0.95\columnwidth]{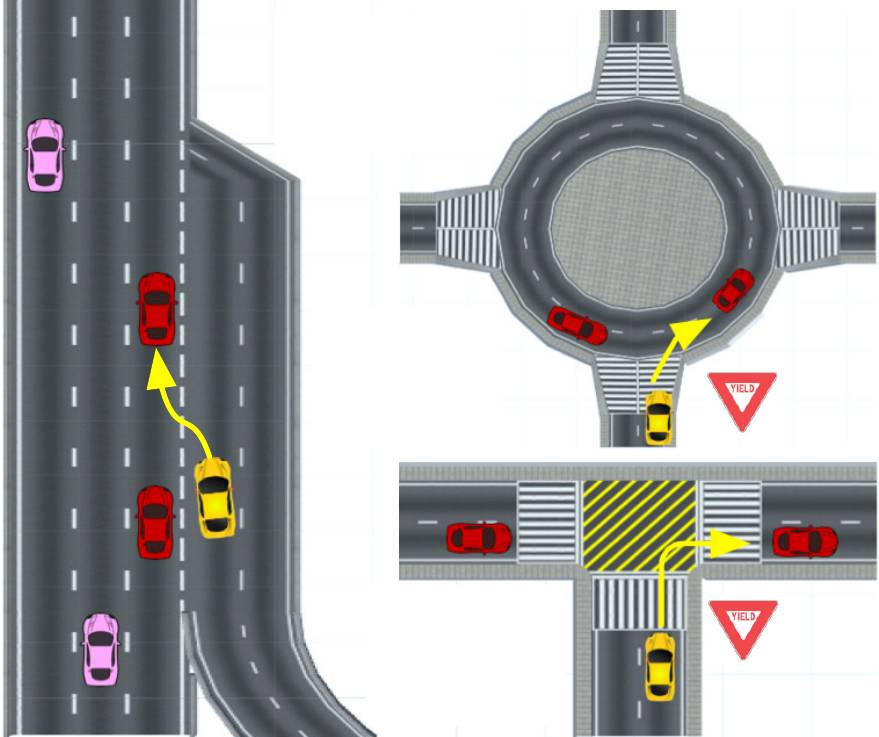}
	\caption{Typical traffic scenarios that lead to strong interaction among vehicles: (a) highway entrance; (b) roundabout; (c) uncontrolled T-intersection. The figures come from the bird-view of several road modules of the driving simulator developed by the authors.}
\end{figure}

A great variety of approaches have been proposed to cope with driver intention and motion prediction problems.
We mainly focus on probabilistic approaches in this paper.
In \cite{A4}, the authors proposed a planning-based prediction approach using an inverse reinforcement learning framework. The road and vehicle states are discretized and the occupancy probability of each cell is obtained through a Markov Decision Process (MDP). However, they did not consider the feasibility of vehicle motions and lateral position distributions within a road lane. 
In \cite{A1}, a probabilistic situation assessment framework was presented in which the vehicle motions are described as several high-level motion patterns and the interaction recognition task is formulated as a classification problem. However, there is no prediction for future scene evolution. 
In \cite{A9}, a collision risk estimation method was introduced in which the driver behavior is modeled by HMM and Gaussian Process (GP) is employed to generate a group of future trajectories of the predicted vehicle. However, the method is only able to predict one car's behavior and the interaction among different entities is not considered.
\cite{A2} presented an intention prediction framework that can estimate the reaction of surrounding vehicles to the ego vehicle behavior, which makes the autonomous car more ``sociable''.
A stochastic reachable set approach proposed in \cite{B1} is based on vehicle kinematic models. However, there is no strong interaction among entities since the right-of-way in the studied scenarios is strict and the vehicle future routes are manually designed which greatly diminishes uncertainties.
In \cite{Wei_metric}, the authors proposed a novel fatality-aware benchmark for probabilistic reaction prediction which evaluate results in more aspects beyond state error.

In this work, we design a hierarchical framework which consists of a situation recognition module providing the probability distribution of each potential situation and propagating the recognition results to a dynamic interactive scene evolution module that generates a set of future trajectories from which the joint distribution of multiple agent states can be obtained. 
This brings the advantages of considering the impact on each entity from its surrounding road participants.
The situation recognition model is based on a two-layer Hidden Markov Model (TLHMM) which is suitable for transfer learning due to the layered representation. This significantly alleviates the need for real-world data since the well-trained model using the simulation data can be finetuned with less naturalistic data. 

The remainder of the paper is organized as follows. Section II presents the previous work related to the proposed approach. Section III introduces the details of the situation recognition model and scene prediction method. In Section IV, a highway ramp merging case study is provided. Section V concludes the paper and suggests future work.

\section{Related Work}
In this section, we introduce the basic concepts of Layered Hidden Markov Model (LHMM) and transfer learning that are closely related to our proposed approach and make a brief overview on the previous work.

\noindent
\textit{\textbf{Layered Hidden Markov Model:}}

Layered Hidden Markov Model (LHMM) has a hierarchical architecture which consists of multiple layers of standard HMM. This layered representation is of great advantages when modeling multi-level activities such as human motions. 
To the best of our knowledge, the concept of LHMM was first proposed in \cite{D1} to recognize human states in an static office environment. 
In \cite{D2}, the method was further applied to human motion intention recognition where the high-level complicated tasks are divided into several low-level primary sub-tasks. The results revealed much better performance for LHMM than a canonical one-layer HMM. 
In \cite{D3}, a Monte Carlo Layered HMM capable of online learning was proposed to predict robot leader's motions, which is used to improve tracking accuracy.

This layer-structured model is suitable for dynamic situation modeling due to the fact that a multi-agent highly interactive process can be decomposed into certain stages, which will be explained in detail in Section III. Moreover, a novel formulation and interpretation for the model is provided.

\noindent
\textit{\textbf{Transfer Learning:}}

The objective of transfer learning is to enhance the learning efficiency and performance of the target tasks through taking advantages of the knowledge learned from the source tasks \cite{E1}. 
In \cite{E4}, the authors proposed a transfer learning method to recognize daily activities of different-aged people in different house settings. A mapping across various sensor network raw data was proposed to extract common meta features among diverse tasks.
In \cite{E5}, Ballan et al. put forward a transfer learning approach where a navigation map learned from the training data and a dynamic Bayesian network are combined to predict traffic participants' motion patterns in novel scenes that are not included in the training phase.
In this paper, our proposed model enables simulation to real world knowledge transfer, which makes up the difficulties of collecting real driving data.

\section{Interactive Situation Recognition and Prediction Approach}

In this section, we first present the generic model architecture for probabilistic situation recognition and prediction in detail. A time efficient training and inference method is then illustrated. At last, we demonstrate how the knowledge learned from simulation data can be transferred to real-world scenarios to reduce training efforts as well as improve recognition performance.
\begin{figure*}
	\centering
	\includegraphics[width=\linewidth]{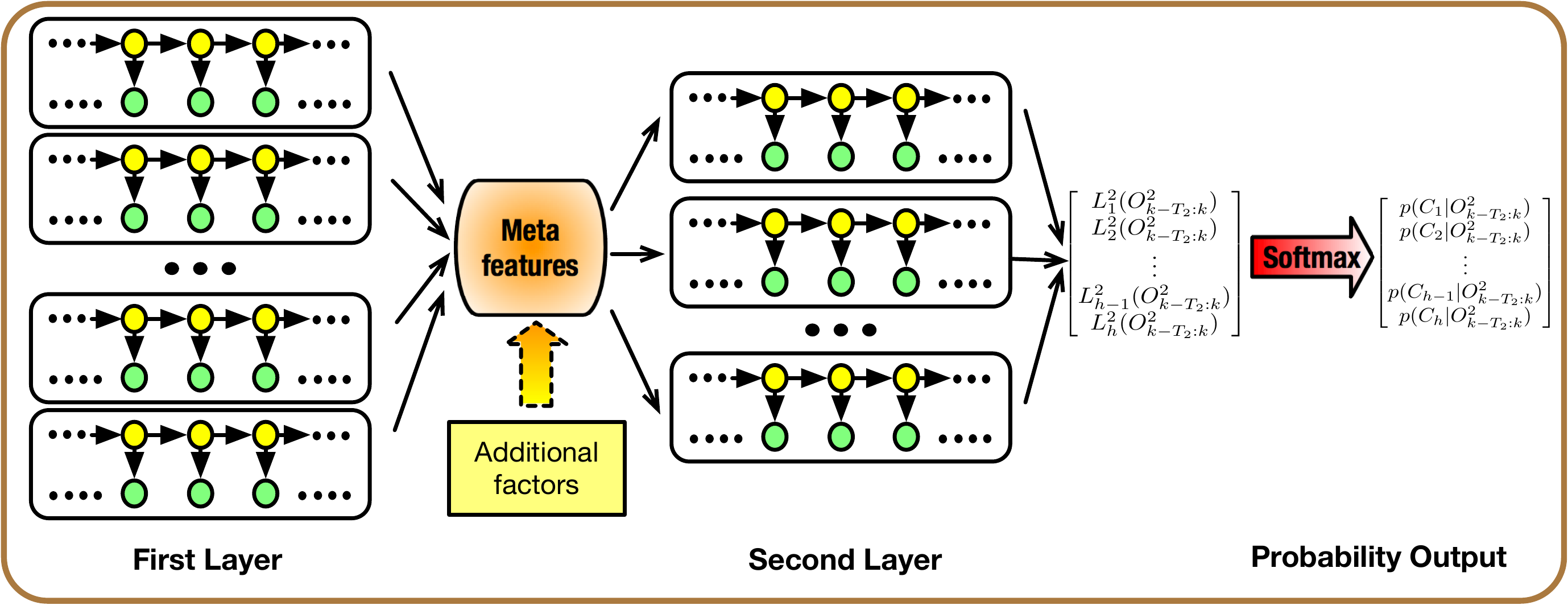}
	\caption{The architecture of the two-layer Hidden Markov Model (TLHMM).}
\end{figure*}

\subsection{Situation Recognition Model}
The recognition model has a two-layer structure which is a cascade of two groups of canonical HMM, which is shown in Fig. 2. The inference output of the first layer is utilized as the observation sequences of the second layer. 
Each HMM in the first layer corresponds to a certain low-level stage of situation evolution while each HMM in the second layer corresponds to a high-level situation which consists of several low-level stages and covers a relatively long period. 
For example, in Fig. 1(b) a complete roundabout merging process in which the vehicle already running in the roundabout (red car) yields the merging vehicle (yellow car) can be divided into three stages: (1) The merging red car decelerates before the yield sign; (2) The yellow car tentatively advances into the conflict zone and the red car decelerates to yield; (3) The yellow car accelerates and the red car begins car-following-like behavior. This specific situation brings three HMM in the first layer and one HMM in the second layer.

The function of the first layer is to extract meta features from the raw feature sequences which can be obtained directly or indirectly from sensor measurement such as road geometry information and traffic participants' states.
The raw observation sequences are sensitive to measurement noise and highly dependent on the environment settings and may follow very diverse distributions, which results in inconsistencies in different scenarios. 
Assume there are $l$ HMMs in the first layer and $\lambda^1_i \ (i=1,2,...,l)$ represents parameters of the $i$-th HMM. We denote the observation at time step $k$ as $O^1_{k}$.
The observation length can be adaptively determined according to particular applications. 
In this paper, we define the observation likelihood obtained by the forward algorithm \cite{HMM} for each first-layer HMM as meta features, which forms a meta feature matrix
\begin{equation*}
\left[
\begin{matrix}
L^1_1(O^1_{k-T_1:k})  & L^1_1(O^1_{k-T_1+1:k})  & \cdots & L^1_1(O^1_{k-1:k})\\
L^1_2(O^1_{k-T_1:k})  & L^1_2(O^1_{k-T_1+1:k})  & \cdots & L^1_2(O^1_{k-1:k})\\
\vdots & \vdots & \ddots & \vdots\\
L^1_l(O^1_{k-T_1:k})  & L^1_l(O^1_{k-T_1+1:k})  & \cdots & L^1_l(O^1_{k-1:k})
\end{matrix}
\right],
\end{equation*}
where $T_1$ is the sliding window horizon of historical information and
$L^1_{i}(O^1_{k-t:k}) = p(O^1_{k-t:k}|\lambda^1_{i}), \ i=1,2,...,l, \ t=1,2,...,T_1$.
This format covers the most comprehensive meta feature set but it can be simplified to a subset in order to reduce computation cost according to the available training data amount, accuracy demands and operational capability.
Considering there may be additional factors such as traffic light state or traffic sign information which are common among different scenarios, one can incorporate such information into the meta feature matrices. 

The second layer serves to obtain the log-likelihood of meta feature sequences given each HMM corresponding to a certain high-level situation. 
Assume there are $h$ HMMs in this layer and $\lambda^2_j \ (j=1,2,...,h)$ represents the parameters of the $j$-th HMM. We denote the observation at time step $k$ as $O^2_{k}$. Note that the observation sliding window horizon $T_2$ can be set different from the first layer due to the decomposition of probabilistic reasoning between layers.
The inference output is a log-likelihood vector which is written as
$\left[
\begin{matrix}
	L^2_1(O^2_{k-T_2:k})  \
	L^2_2(O^2_{k-T_2:k})   \
	\cdots \
	L^2_{h}(O^2_{k-T_2:k}) 
\end{matrix}
\right]^{T}$.
Let $C_j \ (j=1,2,...,h)$ denote the $j$-th situation and assume uniform prior distribution for all the possible situations, then the posterior distribution can be obtained through a softmax layer which gives the probability of each possible situation
\begin{equation*}
p(C_i|O^2_{k-T_2:k}) = e^{L^2_i(O^2_{k-T_2:k}|C_i)} / \sum_{j=1}^{h} e^{L^2_j(O^2_{k-T_2:k}|C_j)}.
\end{equation*}

Due to the decoupling property of the layered representation, the two layers can be trained successively from left to right both with the Baum-Welch algorithm (a.k.a Forward-Backward algorithm) which is a variant of Expectation-Maximization (EM) method \cite{EM}.
The raw feature sequences are used to train the first-layer HMMs and their inference results are used to train the second-layer HMMs. Before training each HMM, we fit a GMM to the training data and set the number of hidden states as the number of GMM components with the highest Bayesian Information Criterion (BIC) score. More details of training process can be found in \cite{D1}. The inference method is also bottom-up in which the observation likelihood of the first layer is propagated to the second layer. The details can be found in \textbf{Algorithm 1}. 

\subsection{Interactive Scene Evolution Prediction}
Learning-based models can be employed to predict a long-term scene evolution. 
First, a prediction model is learned for each potential situation from the corresponding training data.
Then, we can utilize each prediction model to generate a group of trajectories for each objective entities based on the posterior distributions obtained in the recognition module.
In this paper, without loss of generality we adopt a standard GMM to model the joint distribution of states and actions which is similar to the driver behavioral model proposed in the authors' previous work \cite{tracking1} to forecast scene evolution.
\begin{algorithm}[!tbp]
	\caption{Two-Layer HMM Inference Algorithm}
	\begin{algorithmic}[1] 
		\REQUIRE ~~ \\
		1. Test feature sequences ($test\_feature\_seq$) \\
		2. Well-trained HMM of both layers (first: HMM-1; second: HMM-2)
		\ENSURE ~~\\
		The posterior probabilities of each class $probability$.
		\STATE $T \gets len(test\_feature\_seq)$
		\FOR {each first-layer HMM}
		\FOR{$i = 0 \to T-T_1$}
		\STATE $L_1$.append(likelihood($test\_feature\_seq[i:i+T_1]$))
		\ENDFOR
		\ENDFOR
		\STATE $T \gets len(L_1)$
		\FOR{each second-layer HMM}
		\FOR {$j = 0 \to T-T_2$}
		\STATE $L_2$.append(log-likelihood($L_1[j:j+T_2]$))
		\ENDFOR
		\ENDFOR
		\STATE $probability \gets$ Softmax($L_2$)
		\RETURN{$probability$}
	\end{algorithmic}
\end{algorithm}

The Gaussian mixture distribution is a linear combination of multiple Gaussians with the form
$f(\zeta) = \sum_{g=1}^{N} \pi_g \mathcal{N}(\zeta|\mu_g,\Sigma_g)$,
where $\sum_{g=1}^{N} \pi_{g} = 1$, $\mu_{g}$ and $\Sigma_{g}$ are the mean and covariance of the $g$-th Gaussian distribution, and $\mathbf{\zeta}$ is the training dataset.
Fig. 3 shows the conditional dependence among different factors, where $C_i,\ E_k, \ S_k$ and $a_k$ denotes the $i$-th situation, the external environment information, the entities' states and actions at time step $k$. Some proper features can be extracted by processing $E_k$ and $S_k$ and denoted as $SE_k$.
In each training sample, these elements are stacked into a column vector which is denoted as
$\zeta \doteq [\ SE_k \ \vline \ a_k \ ]^T$.
The dimension of each element can be arbitrary. We use the EM algorithm to train the GMM. 
The mean and covariance matrix of the $g$-th component can be decomposed as 
\begin{equation*}
\mu_{g} = [\ \mu_{g,SE} \ | \ \mu_{g,a} \ ]^{T}, \
\Sigma_{g} = \left[ \ \begin{array}{c|c}
\Sigma_{g,SE} & \Sigma_{g,SE,a} \\
\hline
\Sigma_{g,a,SE} & \Sigma_{g,a} \\
\end{array} \ \right].
\end{equation*}
The conditional mean and covariance of actions and the corresponding component weight can be obtained by the same formulas in \cite{tracking1}.
The conditional distribution of action $a_k$ given the current state and environment information $S_k,E_k$ can be obtained and utilized to make short-term or long-term prediction, where the Monte Carlo method is used to sample a sequence of actions.
The conditional state transition distribution can be obtained by
\begin{equation*}
\begin{aligned}
f(SE_{k+1}|SE_k, C_i) &= \int_{a_k} f(SE_{k+1}|SE_k,a_k)f(a_k|SE_k,C_i) \\
f(SE_{k+1}|SE_k) &= \sum_{C_i}f(SE_{k+1}|SE_k,C_i)f(C_i|SE_k)
\end{aligned}
\end{equation*}
According to the first-order Markov assumption, the distribution of predicted trajectories can be calculated by
\begin{equation*}
\begin{split}
f(SE_{k+1}, SE_{k+2},...,SE_{k+T}|SE_k) = f(SE_{k+1}|SE_k)  \\ 
\times f(SE_{k+2}|SE_{k+1})\cdots f(SE_{k+T}|SE_{k+T-1}).
\end{split}
\end{equation*}

\subsection{From Virtual to Real}
The knowledge learned from simulation data can be transferred to real-world applications due to the fact that a set of common meta features among the same category of scenarios can be extracted by the first layer of the proposed model. Since the first layer is highly dependent on traffic environment settings, it must be retrained given a new target task. However, the second layer can be directly transferred or used as initial parameter values for further finetuning with the target task data.
The training process will be much shorter compared with training from scratch and the demands of target task training data can be reduced.
An examplar case is provided in the next section.
\begin{figure}
	\centering
	\includegraphics[width=0.8\columnwidth]{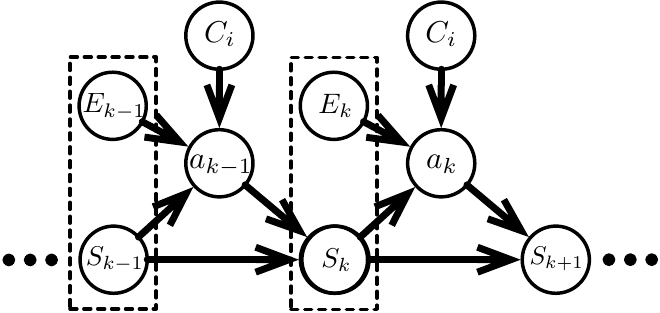}
	\caption{The topological diagram of the scene evolution model for the $i$-th situation.}
\end{figure}

\section{Case Study}
In this section, a typical highway ramp merging scenario is investigated to validate the effectiveness and accuracy of the proposed approach. The data source, experiment details are presented and results are analyzed.

\subsection{Problem Statement}
Among various highway ramps with different road curvature, the schematic diagram can be simplified as Fig. 4. 
\begin{figure}
	\centering
	\includegraphics[width=\columnwidth]{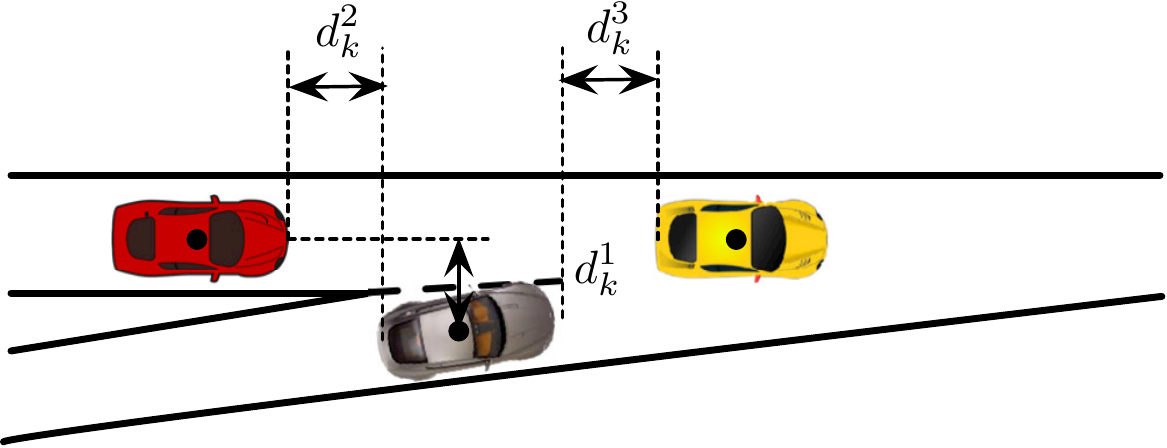}
	\caption{A simplified diagram of highway ramp merging process.}
\end{figure}
The gray car represents the merging vehicle while the red car stands for the car in the main lane with which the merging agent has major interaction. The yellow car is the leading vehicle which affects the available gap for the merging car.
There are two potential interaction outcomes: the red car yields the gray car; or the opposite.
The goal is to obtain the posterior probability of each outcome and the joint distribution of future trajectories of the two entities given a sequence of historical information .

\subsection{Data Collection and Pre-processing}
Our training data and test data come from two sources: virtual simulation and real-world driving dataset. We developed a driving simulator based on Unity3d for data collection and visualization where a steering wheel and two pedals for acceleration and braking are used to collect human inputs for each involved driver. The position, velocity and yaw angle information of all the agents in the scene can be recorded in real time. We collected 128 interaction events in total with balanced distribution of two situations.

We also adopted the Next Generation Simulation (NGSIM) US I-80 Freeway dataset as naturalistic driving data which is available online \cite{NGSIM}. We selected 100 interaction events in total. Since the original data is noisy which results in highly fluctuating velocities and accelerations, we applied an Extended Kalman filter (EKF) to smooth the trajectories. For both simulation data and real-world data, we randomly selected 80\% as the training set and the others were used as test set.

\subsection{Experiment and Results}
There are two potential situations in this scenario: the merging agent cuts in ahead of or behind the main lane agent, corresponding to the two gaps shown in Fig. 4.
A complete merging process can be typically partitioned into four stages:
\begin{itemize}
	\item \textit{Ambiguity}: The two cars advance in their own lanes without strong interaction with each other; 
	\item \textit{Preparation}: At this stage, the two agents are getting closer and the one willing to yield tends to decelerate to avoid collision and provide more space for the other one which may accelerate or keep a constant speed;
	\item \textit{Merging}: At this stage, the merging agent cuts into one of the possible gaps;
	\item \textit{Car following}: The interaction terminates and the yielding one starts car-following-like behavior.
\end{itemize}
The motion patterns and vehicle states in the \textit{Ambiguity} stage have no obvious distinctions between the two situations since the two agents have not made decisions on whether to yield or not. 
However, there exist large differences in the other three stages (e.g. in the merging stage and car following stage, the relative longitudinal positions in the two situations are opposite).

Our experiment contains two parts. The proposed model is first trained and tested with the simulation dataset to demonstrate the recognition and inference performance; then we sample a set of future trajectories to approximate the joint distribution of the two vehicles' states. The well-trained LHMM is also tested on the real-world dataset with or without finetuning to illustrate the transferability. The details are introduced in the following subsections.

\noindent
\textit{\textbf{Virtual Simulation}}:

In this part, we assume that the available gap between the main lane vehicle and the leading vehicle is large enough for the merging vehicle to cut in which implies that the leading car has little influence on the interaction process.

The whole trajectories of each interaction event were divided into four segments corresponding to four stages based on the road information and vehicle states. We trained 7 first-layer HMM individually using the segmented feature sequences. Then we obtained the log-likelihood sequences of both situations from the inference output and fed them to the second layer HMM as training observation sequences. The labels of each HMM are shown in Table I, where HMM-$i$-$j$ refers to the $j$-th HMM in the $i$-th layer.
In order to obtain the conditional distribution of actions $f(a_k|SE_k)$, we trained a Gaussian Mixture Model with the state features and action labels listed in Table II. 
\begin{figure*}[!tbp]
	\centering
	\includegraphics[width=\textwidth,height=9cm]{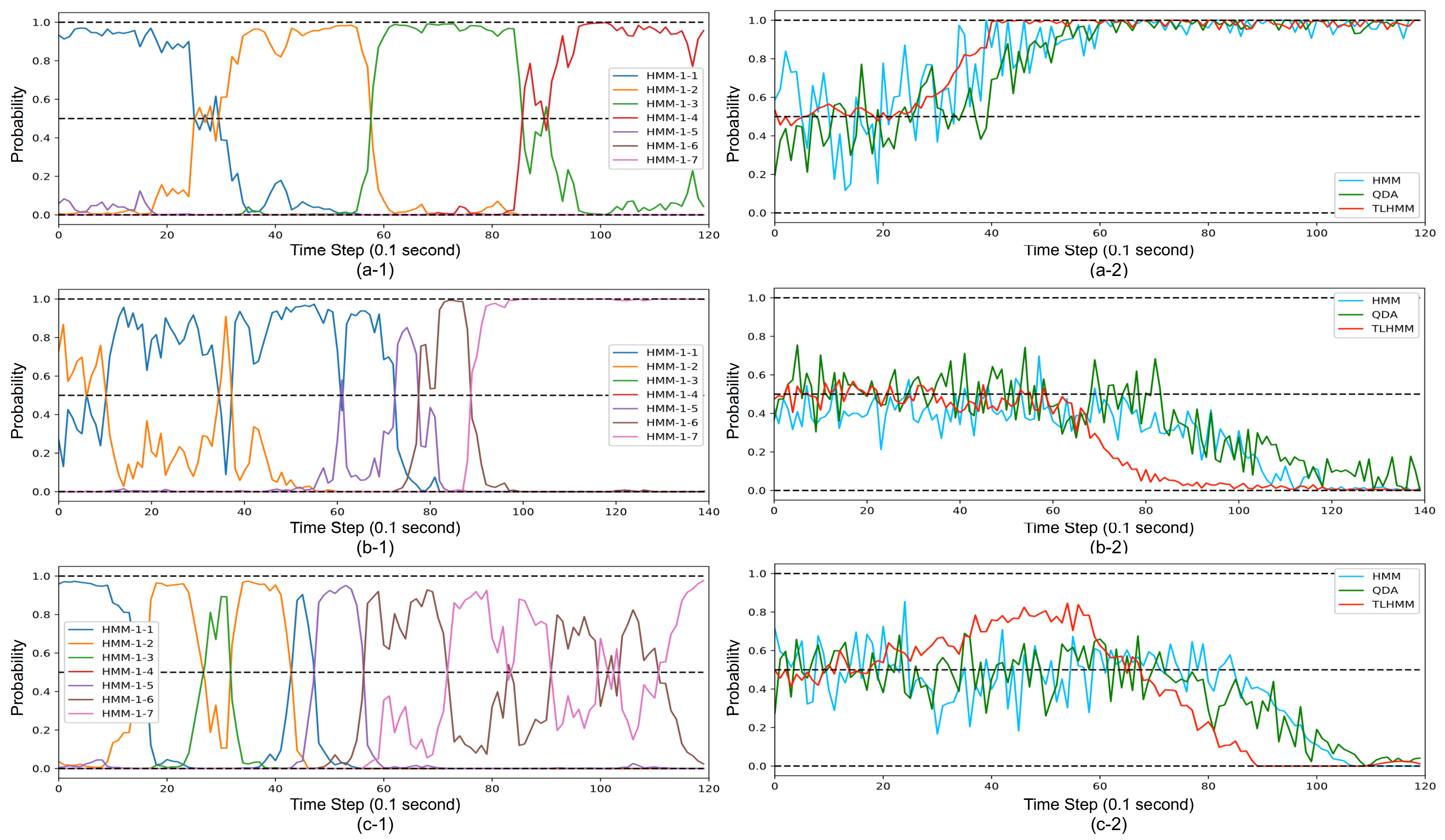}
	\caption{The inference output for three test case: (a) The main lane car yields the merging car; (b) The merging car yields the main lane car; (c) The merging car tends to cut in ahead of the main lane car while the main lane car does not yield. (a-1), (b-1) and (c-1) show the first-layer HMM inference output; and (a-2), (b-2) and (c-2) show the probability output of the second-layer HMM for "\textit{main lane car yields}".}
\end{figure*}
\begin{figure*}[!tbp]
	\centering
	\includegraphics[width=\textwidth,height=6.9cm]{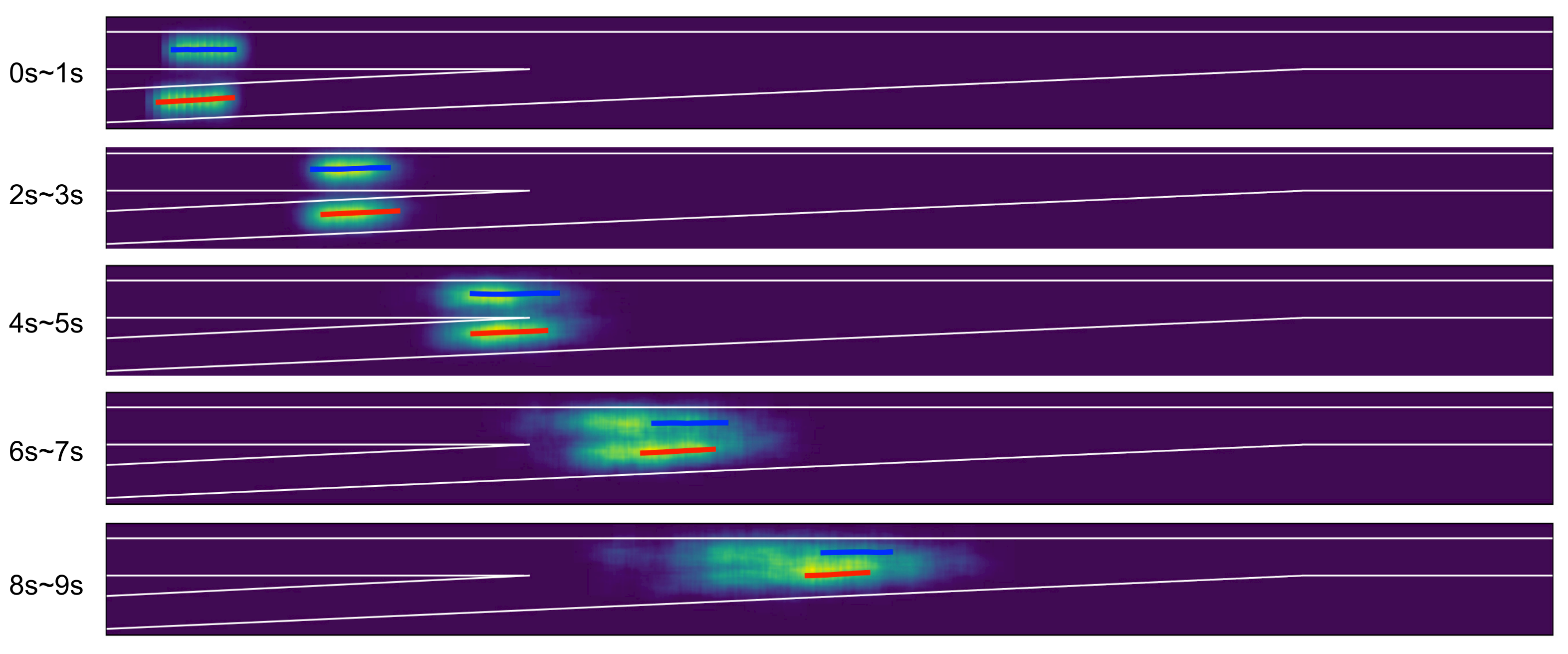}
	\caption{An illustrative heatmap of long-term dynamic scene evolution. It contains predictions of the same case from the same initial state with different propagation length. The red line is the ground truth of the merging vehicle and the blue line is the ground truth of the main lane vehicle.}
\end{figure*}
\begin{table}[!tbp]
	\caption{The Labels of the Two-layer Hidden Markov Model}
	\label{tab:first_layer}
	\begin{center}
		\begin{tabular}{m{1.5cm} m{3.5cm} p{2cm}}
			\toprule
			\midrule
			\textbf{Index} & \textbf{Situation} & \textbf{Stage}\\ 
			\midrule
			HMM-1-1 & Main/Merge lane car yields & Ambiguity\\ 
			HMM-1-2 & Main lane car yields & Preparation\\ 
			HMM-1-3 & Main lane car yields & Merging \\ 
			HMM-1-4 & Main lane car yields & Car following\\ 
			HMM-1-5 & Merge lane car yields & Preparation\\ 
			HMM-1-6 & Merge lane car yields & Merging \\ 
			HMM-1-7 & Merge lane car yields & Car following \\ 
			HMM-2-1 & Main lane car yields & --- \\ 
			HMM-2-2 & Merge lane car yields & --- \\ 
			\bottomrule
		\end{tabular}
	\end{center}
\end{table}
\begin{table}[!tbp]
	\caption{State Features and Action Labels}
	\label{tab:feature}
	\begin{center}
		\begin{tabular}{c m{1cm} p{5cm}}
			\toprule
			\midrule
			&  \textbf{Notations} & \textbf{Descriptions} \\ 
			\midrule
			\multirow{7}*{\shortstack[lb]{State \\ features}}  	
			& $y^1_k$ & Longitudinal position of the main lane car  \\ 
			& $y^2_k$ & Longitudinal position of the merging car  \\ 
			& $d^1_k$ & Lateral distance of two cars \\ 
			& $\dot{y}^1_k$ & Longitudinal velocity of the main lane car \\ 
			& $\dot{y}^2_k$ & Longitudinal velocity of the merging car \\ 
			& $\ddot{y}^1_k$ & Longitudinal acceleration of main lane car \\ 
			& $\ddot{y}^2_k$ & Longitudinal acceleration of main lane car \\ 
			\midrule
			\multirow{6}{*}[0cm]{\shortstack[lb]{Action \\ labels}} 
			& $\Delta x^1_{k \to k+1}$ & Traveled lateral distance of main lane car\\ 
			& $\Delta x^2_{k \to k+1}$ & Traveled lateral distance of merging car \\
			& $\dot{y}^1_{k+1}$ & Longitudinal velocity of main lane car\\ 
			& $\dot{y}^2_{k+1}$ & Longitudinal velocity of merging car\\
			\bottomrule
		\end{tabular}
	\end{center}
\end{table}

Three typical test cases are used to demonstrate the performance of the proposed approach. The inference results of the first-layer HMM are shown in Fig. 5 (a-1), (b-1) and (c-1). We can see that in the first-layer inference output, the observation likelihood of HMM-1-1 is the highest at the early stage for all the test cases, which implies vague decisions of two agents corresponding to the \textit{Ambiguity} stage. Afterwards, the likelihood of HMM-1-2, HMM-1-3 and HMM-1-4 tend to dominate in the ``\textit{main lane car yields}" cases successively while HMM-1-5, HMM-1-6 and HMM-1-7 tend to dominate in the ``\textit{merge lane car yields}" cases successively, which implies that the first-layer HMMs are able to capture different evolution patterns of the raw feature sequences in both situations.

Moreover, we compared the performance of the two-layer HMM with a standard HMM classifier and Quadratic Discriminant Analysis (QDA), which is illustrated in Fig. 5 (a-2), (b-2) and (c-2). We can see that for all the cases our model is able to recognize the true interaction result earliest among the three models. On the other hand, there is much less fluctuation in the probability output of our model than the other models, which implies that our model is more robust to the raw feature fluctuations and observation noise from the sensor measurement. The reason is that while the inference output of standard HMM and QDA are very sensitive to raw feature evolution, our model are sensitive to meta feature evolution, which reduces the effects of raw feature noise.

An occupancy heatmap is provided in Fig. 6 to illustrate the predicted distribution of future trajectories generated by the GMM. The ground truth trajectory is located near the mean of the distribution, which implies that our model is able to make accurate long-term prediction. 

\noindent
\textit{\textbf{Simulation to Real World Transfer}}:

In this part, we investigate the capability of knowledge transfer of the proposed model. In the NGSIM dataset, since the distances among vehicles are relatively small and the traffic speed is low, the influence of the leading vehicle cannot be ignored. Therefore, we added the position and velocity information of the leading vehicle into the raw feature sequences.
Since the road geometry and vehicle speed range in the NGSIM dataset are different from the simulation data, the first-layer HMM has to be retrained. For the second-layer HMM, three model setups were evaluated: 1) directly employ the well-trained second layer HMM without finetuning the parameters; 2) employ the well-trained parameters as the initial values and finetune the second-layer HMM; and 3) train the two-layer HMM from scratch using the NGSIM dataset. 
It took 112 iterations in average for the EM algorithm to converge when training the model from scratch but only 35 iterations when finetuning the pre-trained model.
The inference results of two test cases are demonstrated in Fig. 7. We can see that the transferred model with finetuning has comparable performance to the model trained from scratch with much less computation cost.
\begin{figure}[!tbp]
	\centering
	\includegraphics[width=\columnwidth,height=7.2cm]{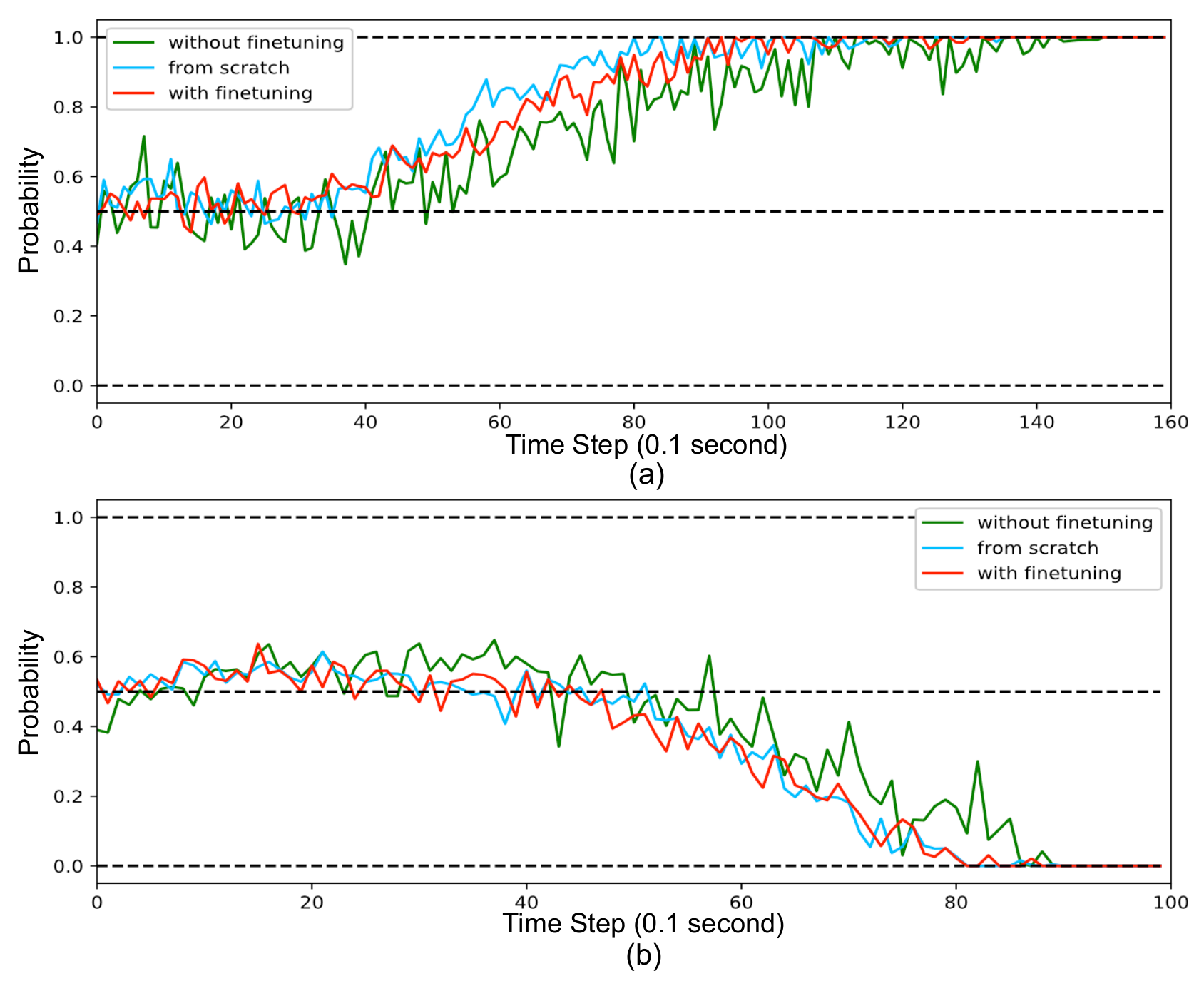}
	\caption{The comparison of inference results among three model setups. (a) a \textit{"main lane car yields"} case; (b) a \textit{"merging car yields"} case.}
\end{figure}

\section{CONCLUSIONS}
In this paper, a probabilistic interactive situation recognition and prediction approach was proposed which consists of two hierarchical modules, where a two-layer Hidden Markov Model is employed to obtain the situation distribution and a learning-based behavioral model is used to predict the interactive scene evolution. Due to the advantages of layered representation, the proposed model is suitable for transfer learning which enables knowledge transfer from simulation to the real world.
A case study of highway ramp merging scenario was conducted to demonstrate the prediction performance as well as the practicability of transfer learning.
For future work, the proposed method will be applied to more challenging scenarios such as intersections and round abouts. Moreover, we will design more complicated probabilistic graphical models to make long-term predictions.


\bibliographystyle{IEEEtran}
\bibliography{reference}

\end{document}